\documentclass[10pt,twocolumn,letterpaper]{article}

\usepackage{cvpr}              

\usepackage[dvipsnames]{xcolor}

\usepackage{amsmath}
\usepackage{multirow}
\usepackage{makecell}

\definecolor{cvprblue}{rgb}{0.21,0.49,0.74}
\usepackage[pagebackref,breaklinks,colorlinks,citecolor=cvprblue]{hyperref}

\def\paperID{****} 
\def\confName{CVPR}
\def\confYear{2024}

\title{Detecting As Labeling: \\ Rethinking LiDAR-camera Fusion in 3D Object Detection}

\author{Junjie Huang\thanks{Corresponding author. junjie.huang@ieee.org},
    Yun Ye,
    Zhujin Liang,
    Yi Shan,
    Dalong Du \\
    Phigent Robotics, Beijing, China}

\begin{document}
\maketitle

\begin{abstract}
3D object Detection with LiDAR-camera encounters overfitting in algorithm development which is derived from the violation of some fundamental rules. We refer to the data annotation in dataset construction for theory complementing and argue that the regression task prediction should not involve the feature from the camera branch. By following the cutting-edge perspective of 'Detecting As Labeling', we propose a novel paradigm dubbed DAL. With the most classical elementary algorithms, a simple predicting pipeline is constructed by imitating the data annotation process. Then we train it in the simplest way to minimize its dependency and strengthen its portability. Though simple in construction and training, the proposed DAL paradigm not only substantially pushes the performance boundary but also provides a superior trade-off between speed and accuracy among all existing methods. With comprehensive superiority, DAL is an ideal baseline for both future work development and practical deployment. The code has been released to facilitate future work on \url{https://github.com/HuangJunJie2017/BEVDet}.
\end{abstract}

\section{Introduction}
With superior robustness, 3D object detection with LiDAR-camera fusion plays an important role in robotics. Fueled by the vision of autonomous driving, many efforts \cite{CMT, UniTR, TransFusion, SparseFusion, BEVFusion, BEVFusion-Adalab, DeepIter, UVTR} have been devoted to this topic in the past few years. Nevertheless, some fundamental rules are violated in all these existing methods, which makes them struggle with overfitting. As a remedy, complicated training strategies are applied along with the existing algorithms as listed in Tab~\ref{tab:trainingpipeline}. They use multiple training stages or customized learning rate policies to alleviate this problem. However, what they achieve is just a local optimum that will trap future work to a large extent. Besides, additional dependence (\texttt{e.g.} pre-training on other datasets) also gives rise to extra cost and uncertainty in practice.

\begin{figure}
	\setlength{\abovecaptionskip}{0.cm}
    \begin{center}
        \includegraphics[width=0.95\hsize]{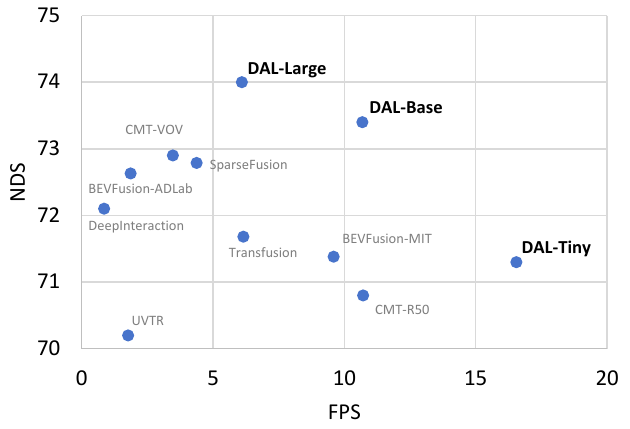}
    \end{center}
   \caption{Trade off between performance and inference speed on the nuScenes \cite{nuScenes} val set. For a fair comparison, we applied the same accelerated voxelization module to all algorithms and reported their inference speed after acceleration. }
    \label{fig:fps-nds}
\vspace{-0.2cm}
\end{figure}

\begin{table*}[t]
\footnotesize
\begin{center}
\begin{tabular}{c|c|c|c|c|c}

\hline
Methods             &Camera Pre.                                   &LiDAR Pre.   & Custom Learning Rate                                      & Epochs  & NDS   \\
\hline
UVTR\cite{UVTR}   &ImageNet $\rightarrow$ nuScenes                      & -                 & ALL Except Head $\times$ 0.1                              & 20        & 70.4  \\
BEVFusion\cite{BEVFusion-Adalab}   &ImageNet $\rightarrow$ nuScenes     & TransFusion-L     & -                                                         & 10        & 72.1  \\
BEVFusion(MIT)\cite{BEVFusion}   &ImageNet $\rightarrow$ nuImages       & TransFusion-L     & -                                                         & 6         & 71.4  \\
TransFusion\cite{DeepIter}      &ImageNet $\rightarrow$ COCO            & TransFusion-L     & -                                                         & 6         & 71.7  \\
DeepIteraction\cite{DeepIter}  &ImageNet $\rightarrow$ COCO $\rightarrow$ nuImages          & TransFusion-L& -                                          & 6         & 72.6  \\
CMT-VOV\cite{CMT}   &ImageNet $\rightarrow$ DD3D $\rightarrow$ nuScenes & -                 & Image Backbone $\times$ 0.01 / Image Neck $\times$ 0.1    & 20        & 72.9  \\
SparseFusion\cite{SparseFusion}   &ImageNet $\rightarrow$ nuImages      & TransFusion-L     & Image Backbone and Neck $\times$ 0.1/ Freeze LiDAR        & 6         & 72.8  \\
UniTR\cite{UniTR}   &ImageNet $\rightarrow$ nuImages      & -     & -        & 10         & 73.3  \\
\hline
DAL-Large                 &ImageNet                                     & -                 & -                                                   & 20        & \textbf{74.0}  \\
\hline
\end{tabular}
\end{center}
\vspace{-0.5cm}
\caption{Comparing the DAL training pipeline with the existing methods. Existing methods use multiple pre-training stages or complicated learning rate policies to prevent the fusion model from overfitting. The proposed DAL paradigm only pre-trains the image backbone on ImageNet but performs better.}
\label{tab:trainingpipeline}
\vspace{-0.2cm}
\end{table*}

We rethink LiDAR-camera fusion in 3D object detection by referring to the data annotation phase in dataset construction \cite{nuScenes}. In this phase, annotations are generated in two steps: candidates are searched for in images and point cloud at first, and the 3D bounding boxes are annotated according to the LiDAR points of each instance (\texttt{i.e.} tightly includes all LiDAR points). To guarantee high quality in labeling, some rules should be strictly obeyed by the annotators:
\begin{itemize}
    \item[A.] Images should be incorporated with point cloud to search all possible candidates and determine their categories.
    \item[B.] The 3D bounding box of each instance should be generated only according to the point cloud when the point cloud is sufficiently complete for locating the edges of the bounding box.
\end{itemize}
Rule.B demands a priority disparity between the point cloud and the image in annotating the 3D bounding boxes. This is derived from the different robustness of these two patterns in this sub-task: The point cloud is an infallible ruler while the vision is merely an experienced gambler. The ill-posed monocular depth estimation problem makes the vision inevitably not robust on this topic. Violating Rule.B and intuitively involving the image features in predicting the regression targets makes the existing methods fall into the dilemma of overfitting.

Following the perspective of 'Detecting As Labeling', we construct a demonstrative framework, dubbed DAL, to reveal the value of this theory to this problem. We only use the most classical elements to build its elegant predicting pipeline. Then we train it in the simplest manner to minimize its dependency and strengthen its portability. In addition, we notice that the planning task in autonomous driving takes advantage of the movable objects' velocity for active safety. However, the practical data always has an extremely imbalanced velocity distribution, which degenerates the perception algorithm's performance on this topic. This inspires us to develop a velocity augmentation strategy to alleviate this problem.

As a result, though simple in construction and training, the proposed DAL paradigm not only substantially pushes the performance boundary (e.g. 74.0 NDS on the nuScenes \texttt{val} set and 74.8 NDS on the nuScenes \texttt{test} set) but also provides a superior trade-off between speed and accuracy among all existing methods. The main contributions of this paper can be summarized as follows:
\begin{itemize}
  \item [1.] We propose a cutting-edge perspective of 'Detecting As Labeling' for LiDAR-camera fusion in 3D object detection. It is a good patch to existing methods and also should be a fundamental rule obeyed by future works.

  \item [2.] We follow the perspective of 'Detecting As Labeling' to build a robust paradigm dubbed DAL. DAL is the first LiDAR-Camera fusion paradigm with an extremely elegant training pipeline. Besides, it substantially pushes the performance boundary of this problem alone with a superior trade-off between inference latency and accuracy. With comprehensive superiority, DAL is an ideal baseline for both future work development and practical usage.

  \item [3.] We point out the inevitable imbalance problem of velocity distribution and propose instance-level velocity augmentation to alleviate this problem.
\end{itemize}

\begin{figure*}
	\setlength{\abovecaptionskip}{0.cm}
    \begin{center}
        \includegraphics[width=0.90\hsize]{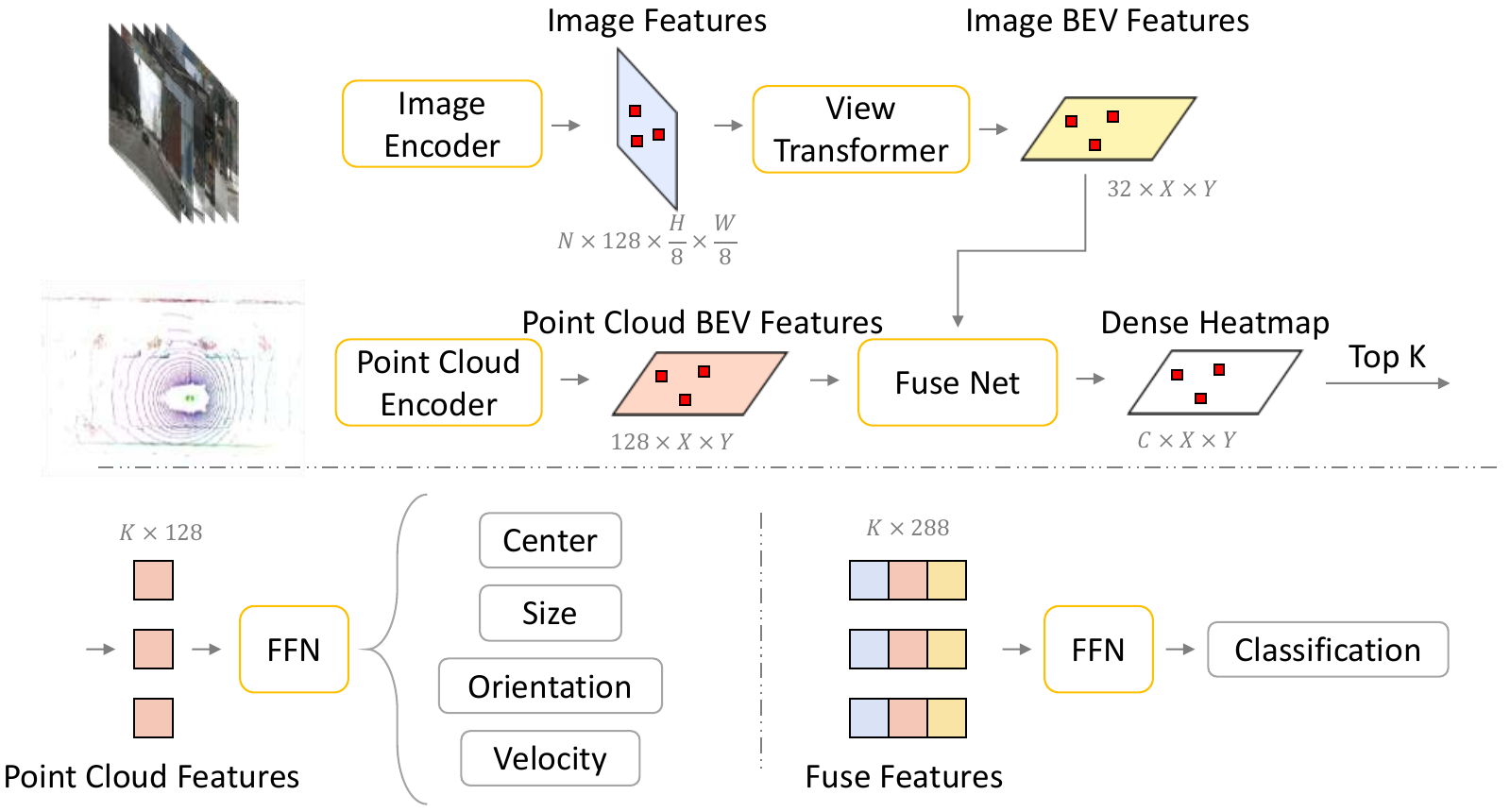}
    \end{center}
   \caption{The prediction pipeline of DAL paradigm. BEV features from the image and point cloud are fused to generate the dense heatmap. Top $K$ proposals and their point cloud features are extracted for regression task prediction. Features fused with the image feature, the image BEV feature, and the point cloud BEV feature are used for category prediction. The sparse image feature is extracted according to the corresponding predicted center of each proposal.}
    \label{fig:pipeline}
\vspace{-0.2cm}
\end{figure*}

\section{Related Work}
\label{sec:RW}
As complementary signals, LiDAR-camera fusion offers appealing performance in 3D object detection. It has been a long literature for engineers developing various works on this topic. The previous works can be roughly classified into three paradigms: Early works \cite{MVXNet, MVP, SFD, VSC, PointPainting, PointAugmenting} prefer to strengthen the point cloud in aspects of amount complement and feature enhancement. Recently, inspired by the successes of camera-only 3D object detection in BEV \cite{BEVDet, BEVDet4D, BEVDepth, SOLOFusion}, some works \cite{FSC, UVTR, BEVFusion, BEVFusion-Adalab} focus on feature fusion in a unified space. Following the mechanism of attention, some other works \cite{TransFusion,ObjectFusion,CMT,UniTR,SparseFusion} transfer all modalities into tokens and always predict targets in a sparse paradigm \cite{DETR}. They update the features of sparse proposals by applying attention calculation with both the image features and the point cloud features. DAL focuses on the rectification of fundamental theory. It absorbs some excellent features from the previous works to construct itself. And in turn, it is a promising direction for revising the existing works. Besides, as an impressive milestone, it is also an excellent baseline for future work and practical usage.

\section{Detecting As Labeling}
\label{sec:MT}

\subsection{Network}
Following the perspective of 'Detecting As Labeling', we construct a predicting pipeline by imitating the data annotation process as illustrated in Fig.~\ref{fig:pipeline}. The proposed pipeline follows the dense-to-sparse paradigm \cite{TransFusion}. The dense perception stage focuses on feature encoding and candidate generating. The features from the images $F_I \in \mathbb{R}^{N\times 128\times H\times W}$ and the point cloud $F_P \in \mathbb{R}^{128\times X\times Y}$ are extracted with an image encoder and a point cloud encoder separately. $N$ describes the number of views. $H\times W$ describes the size of the feature in the image view. $X\times Y$ describes the size of the feature in Bird-Eye-View(BEV). The feature encoders have a classical structure with a backbone (\texttt{e.g.} ResNet\cite{ResNet} and VoxelNet \cite{VoxelNet}) and a neck (\texttt{e.g.} FPN\cite{FPN} and SECOND \cite{SECOND}. Then the image feature is transformed from the image view to BEV with the classical view transform algorithm Lift-Splat-Shot(LSS) \cite{LSS}. We just fuse the dense image BEV feature and point cloud BEV feature by concatenation and predicting the dense heatmap $H \in \mathbb{R}^{C\times X\times Y}$ by applying two extra residual blocks \cite{ResNet}. $C$ describes the number of categories. Finally, $K$ candidates with leading predicted scores in the dense heatmap are selected. In this way, we imitate the candidate-generating process in data annotation. Features from both image and point cloud are used in this process for a complete set of candidates.

In the sparse perception stage, the point cloud feature of each candidate is first gathered according to its coordinates in the dense heatmap. Then the regression targets (\texttt{e.g.} center, size, orientation, and velocity) are predicted with a simple Feed-Forward Network (FFN). No image feature is involved in this process to prevent the overfitting problem. Finally, we fuse the image feature, the image BEV feature, and the point cloud BEV feature to generate a fused feature for category prediction. The section from the image BEV feature is extracted according to the candidate's coordinates in the dense heatmap while the section from the image feature is extracted according to the predicted object center.

The prediction pipeline of DAL inherited most structural design from BEVFusion \cite{BEVFusion} except for some key modifications. First of all, the point cloud BEV feature and the image BEV feature are fused after the dense BEV encoder while BEVFusion fuses them before. We postpone the fusion to retain the regression ability of the LiDAR branch to a maximum extent. Then, the attention between the sparse instance and the BEV feature is removed as we found it unnecessary. Finally, the regression tasks are predicted with the point cloud features only while BEVFusion \cite{BEVFusion} uses the fused features.

\subsection{Training}
\label{sec:MT-train}
As we have assigned tasks with proper modalities in constructing the predicting pipeline, we merely need to load the parameter of the image backbone pre-trained on ImageNet \cite{ImageNet} like most classical vision tasks \cite{COCO, ADE20K}. Then we train DAL in an end-to-end pattern with only one stage. Only data from the target dataset nuScenes \cite{nuScenes} is used. In this way, we train the DAL models in the most elegant manner which is rare in the literature.

As an example, DAL shares the same design of targets and losses with TransFusion \cite{TransFusion} and BEVFusion \cite{BEVFusion}. Other than that, we add an auxiliary classification head upon the image features to strengthen the image branch in searching candidates and discriminating different categories. This is important for DAL as the supervisions from both dense and sparse perception phases in the 3D object detection head are defective. Specifically, in the dense perception phase, the image features are adjusted according to the predicted depth score in the view transformation. So is the gradient in back-propagation. A defective predicted depth score is inevitable and so is the supervision. In the sparse perception phase, instead of the image features of all annotated targets, only those of predicted instances are involved in the loss calculation. An auxiliary classification head with supervision from all annotated targets can resolve the aforementioned problems and strengthen the image branch to a certain extent. In practice, the gravity centers of annotated targets are used to extract the sparse feature of each annotated target. Then classification is conducted on the sparse feature with another FFN, and the loss is calculated just the same as the classification task in 3D object detection head \cite{TransFusion}. Without re-weighting, we directly add the auxiliary loss to the existing losses:
\begin{equation}
    L_{\text{DAL}} = L_{\text{aux}} + L_{\text{Transfusion}}
\end{equation}

The deprecation of image features in regression task prediction not only prevents the inevitable performance degeneration but also enables a wider range of data augmentation in the image space. We take the resize augmentation as an example for explanation. Camera-based 3D object detection predicts the size of the target according to its size in the image view. When the image is randomly resized, to maintain the consistency between the image features and the predicted targets, adjustment is needed to be conducted on the predicted target accordingly. Then is the point cloud in a chain reaction in 3D object detection with LiDAR-camera fusion. So existing methods always use a small range of data augmentation in the image space. This, as a result, keeps them away from the benefit of large-scale data augmentation in the image space like that in most image 2D tasks (\textit{e.g.} classification \cite{ImageNet}, detection\cite{COCO}, segmentation\cite{COCO}.).

\begin{figure}[t]
	\setlength{\abovecaptionskip}{0.cm}
    \begin{center}
        \includegraphics[width=1.0\hsize]{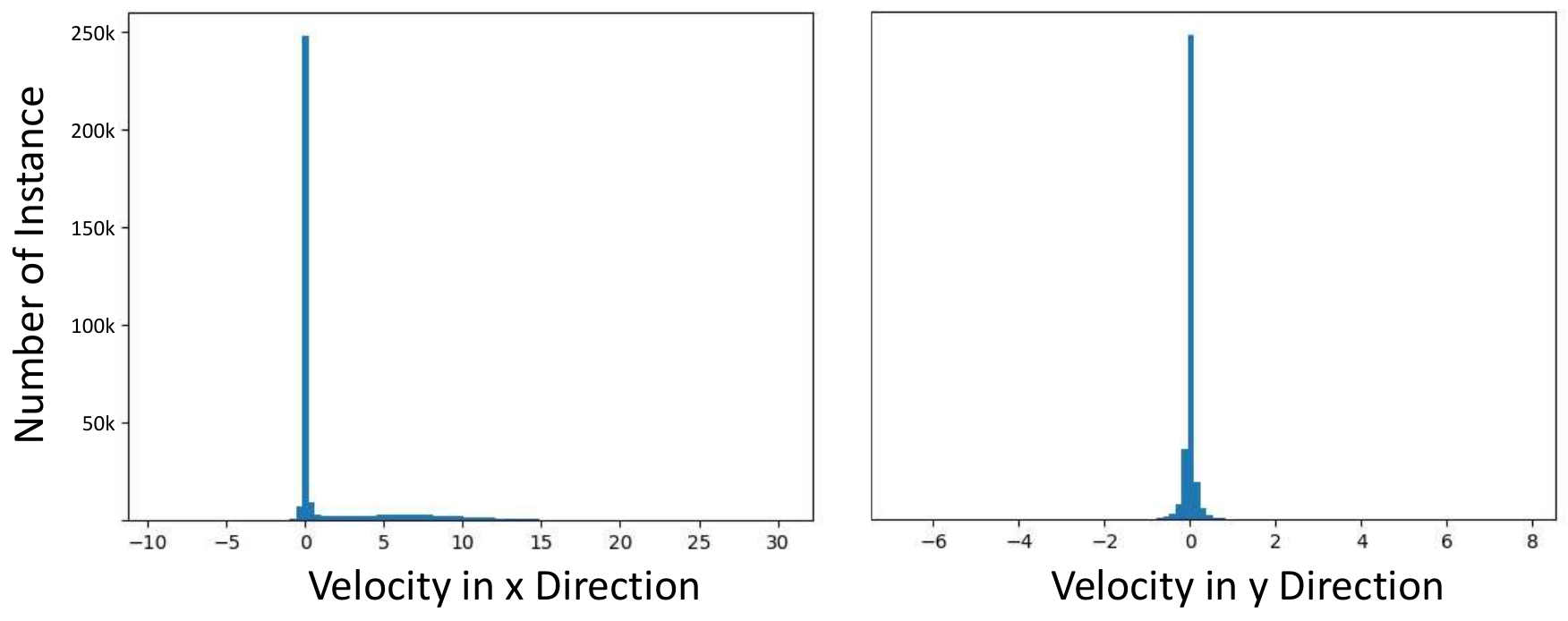}
    \end{center}
   \caption{The velocity distribution of the car category in nuScenes \cite{nuScenes} train set. }
    \label{fig:velocity}
\end{figure}

\begin{figure}[t]
    \setlength{\abovecaptionskip}{0.cm}
    \begin{center}
    \includegraphics[width=1.0\hsize]{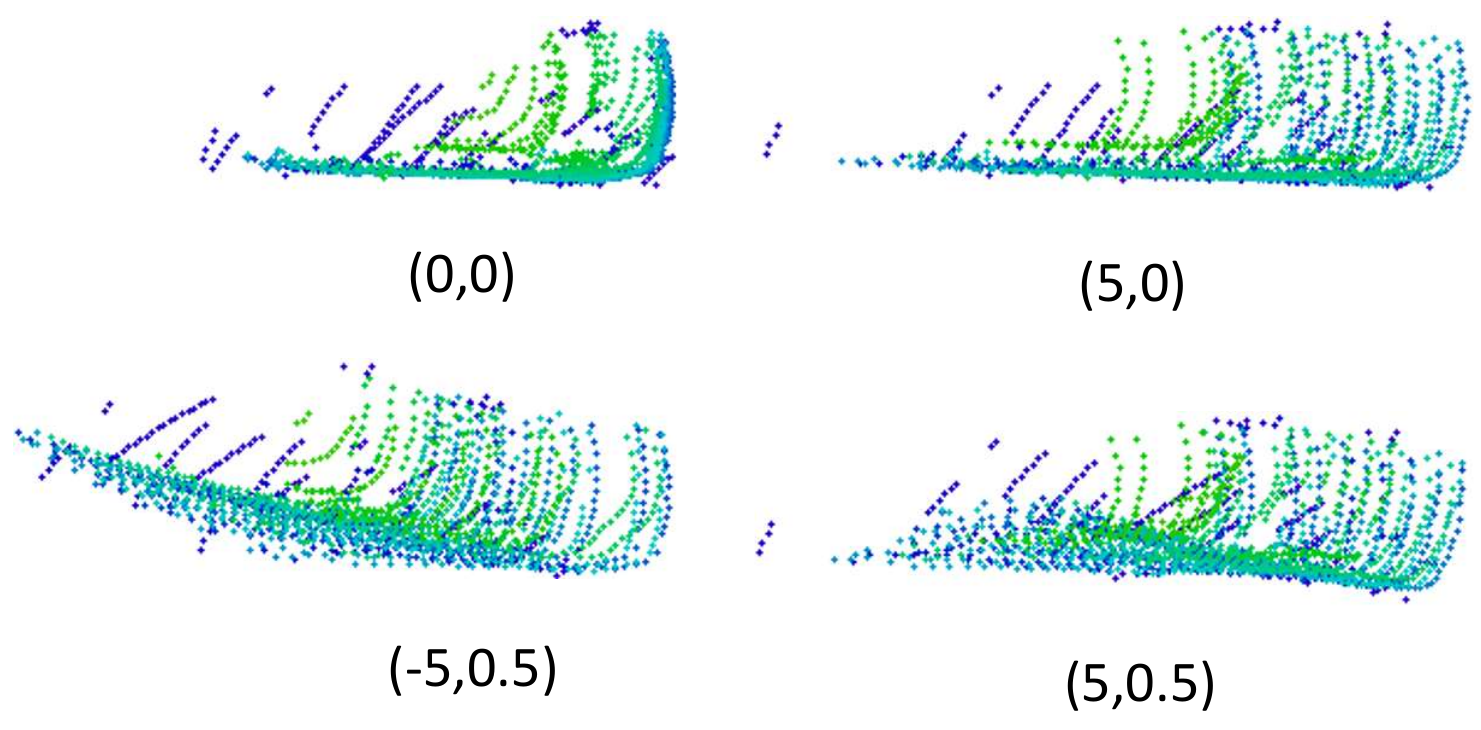}
    \end{center}
   \caption{Augmentation of the same static object with different pre-defined velocities (\texttt{i.e.} $(v_x,v_y)$).}
    \label{fig:velocity_aug}
\vspace{-0.2cm}
\end{figure}

Last but not least, we observe extremely imbalanced velocity distribution of the training data. As illustrated in Fig.~\ref{fig:velocity}, most instances of car category in nuScenes \cite{nuScenes} train set is static. To adjust the distribution, some static objects are randomly selected and their point clouds are adjusted according to a predefined velocity as illustrated in Fig.~\ref{fig:velocity_aug}. We conduct velocity augmentation on static objects only as a full set of its points from multiple LiDAR frames can be handily discriminated with its annotated bounding box.

\begin{table}
\footnotesize
\begin{center}
\begin{tabular}{c|c|c|c|c}
\hline
 Component            & Ablation          & Tiny              & Base              & Large \\
\hline
Image Resolution    & $256\text{x}704$    & $192\text{x}544$    & $256\text{x}704$    & $384\text{x}1056$ \\
\hline
Image Backbone      & R50 \cite{ResNet} & R18 \cite{ResNet} & R18 \cite{ResNet} & R50 \cite{ResNet}\\
\hline
Image Neck          & FPN \cite{FPN}    & FPN \cite{FPN}    & FPN \cite{FPN}    & FPN \cite{FPN}   \\
\hline
\multirow{2}{*}{\makecell{Voxel Resolution \\(Meter)}} & \multirow{2}{*}{0.100} & \multirow{2}{*}{0.100} & \multirow{2}{*}{0.075} & \multirow{2}{*}{0.050} \\
&&&&\\
\hline
\multirow{2}{*}{\makecell{Sparse Encoder\\(Base Channels)}}     & \multirow{2}{*}{16}    & \multirow{2}{*}{16}    & \multirow{2}{*}{24}    & \multirow{2}{*}{32}   \\
&&&&\\
\hline
\multirow{3}{*}{\makecell{Dense Encoder\\ (Stride-Blocks\\-Channels)}} & \multirow{3}{*}{\makecell{1-5-128\\2-5-256}} & \multirow{3}{*}{\makecell{1-5-128\\2-5-256}} & \multirow{3}{*}{\makecell{1-8-192\\2-8-384}} &1-3-128\\
& &  &  &2-3-256\\
&       &       &       &2-3-256\\
\hline
BEV Neck        & SEC.\cite{SECOND} & SEC.\cite{SECOND}& SEC.\cite{SECOND}& SEC.\cite{SECOND}\\
\hline
\end{tabular}
\end{center}
\vspace{-0.5cm}
\caption{Details of the typical DAL predicting pipelines.}
\label{tab:network}
\vspace{-0.2cm}
\end{table}

\begin{table*}
\footnotesize
\begin{center}
\begin{tabular}{l|c|cccccc|c}
\hline
Method                          &Present at   &mATE   &mASE            &mAOE                            &mAVE       &mAAE    & mAP   & NDS \\
\hline
PointPainting \cite{PPaint}     &CVPR'20     &38.0       &26.0        &54.1      & 29.3      &13.1       &54.1       &61.0 \\
PointAugmenting\cite{PAug}      &CVPR'21     &25.3       &23.5        &35.4      & 26.6      &12.3       &66.8       &71.1 \\
UVTR \cite{UVTR}                 &NeurIPS'22     &30.6       &24.5        &35.1      & 22.5      &12.4       &67.1       &71.1 \\
FusionPainting \cite{FPaint}    &ITSC'21     &25.6       &23.6        &34.6      & 27.4      &13.2       &68.1       &71.6 \\
TransFusion \cite{TransFusion}  &CVPR'22     &25.9       &24.3        &32.9      & 28.8      &12.7       &68.9       &71.7 \\
BEVFusion(MIT) \cite{BEVFusion} &ICRA'23     &26.1       &23.9        &32.9      & 26.0      &13.4       &70.2       &72.9 \\
BEVFusion(ADLab) \cite{BEVFusion-Adalab} &NeurIPS'22  &25.0 &24.0        &35.9      & 25.4      &13.2       &71.3       &73.3 \\
ObjectFusion \cite{ObjectFusion} &ICCV'23  &- &-        &-      & -      &-       &71.0       &73.3 \\
DeepIteraction \cite{DeepIter}  &NeurIPS'22     &25.7       &24.0        &32.5      & 24.5      &12.8       &70.8       &73.4 \\
SparseFusion \cite{SparseFusion}&ICCV'23     &25.8       &24.3        &32.9      & 26.5      &13.1       &72.0       &73.8 \\
CMT \cite{CMT}                  &ICCV'23     &27.9       &23.5        &30.8      & 25.9      &11.2       &72.0       &74.1 \\
UniTR \cite{UniTR}                  &ICCV'23     &24.1       &22.9       &25.6      & 24.0      &13.1       &70.9      &74.5 \\
\textbf{DAL}                    &       &25.3       &23.9        &33.4      & 17.4      &12.0       &72.0       &\textbf{74.8} \\
\hline
\end{tabular}
\end{center}
\vspace{-0.5cm}
\caption{Comparisons on nuScenes \texttt{test} set. All present methods use a single model without any test time augmentation.}
\label{tab:test}
\end{table*}

\begin{table*}
\footnotesize
\begin{center}
\begin{tabular}{l|l|l|c|r|cccccc}
\hline
Method                          &LiDAR            &Camera                            &\#Params(M)& FPS & mAP   & NDS \\
\hline
CMT-R50\cite{CMT}               &0100VoxelNet     &$320\times800$-R50               & 40.98     &10.72(14.2${^\dagger}$)  &67.9   & 70.8  \\
\textbf{DAL-Tiny}               &0100VoxelNet     &$192\times544$-R18               & 21.21     &16.55  &67.4   &71.3 \\
\hline
UVTR\cite{UVTR}                 &0075VoxelNet+    &$900\times1600$-R101-DCN \cite{DCN}        & 88.85     &1.77   &65.4   & 70.4  \\
BEVFusion(MIT) \cite{BEVFusion} &0075VoxelNet     &$256\times704$-STTiny \cite{Swin} & 40.84  &9.58   &68.5   & 71.4  \\
TransFusion \cite{TransFusion}  &0075VoxelNet     &$448\times800$-R50               & 37.03     &6.51   &68.9   & 71.7  \\
ObjectFusion \cite{ObjectFusion}&0075VoxelNet     &$256\times704$-STTiny \cite{Swin}& -     &-   &69.8   & 72.3  \\
DeepIteraction \cite{DeepIter}  &0075VoxelNet     &$448\times800$-R50               & 57.90     &1.86   &69.9   & 72.6  \\
SparseFusion \cite{SparseFusion}&0075VoxelNet+    &$448\times800$-R50               & 40.16     &4.38   &70.5   & 72.8   \\
CMT-VOV \cite{CMT}              &0075VoxelNet     &$640\times1600$-VOVNet  & 86.67     &3.48(6.0${^\dagger}$)   &70.3   & 72.9  \\
UniTR \cite{UniTR}              &0030DSVT \cite{DSVT} &$256\times704$-DSVT          & 15.56     &(9.3${^\dagger}$)   &70.5   & 73.3  \\
\textbf{DAL-Base}               &0075VoxelNet+    &$256\times704$-R18               & 35.06     &10.69  &70.0   &73.4 \\
\hline
\textbf{DAL-Large}              &0050VoxelNet+    &$384\times1056$-R50               &47.77      & 6.10  &\textbf{71.5}   &\textbf{74.0} \\
\hline
\end{tabular}
\end{center}
\vspace{-0.5cm}
\caption{Comparisons on nuScenes \texttt{val} set. FPS is measured on a 3090 GPU by default. + means modification is applied as listed in Tab.~\ref{tab:network}. ${^\dagger}$ denotes the inference speed on A100 GPU referred from the original paper.}
\label{tab:val}
\vspace{-0.2cm}
\end{table*}

\section{Experiments}
\label{sec:EP}

\subsection{Implementation Details}

\paragraph{Dataset} We conduct comprehensive experiments on the large-scale benchmark nuScenes \cite{nuScenes}. NuScenes is the up-to-date popular benchmark for verifying many outdoor tasks like 3D object detection \cite{DETR3D, DD3D, BEVDet, BEVDet4D, CenterPoint, BEVFusion}, occupancy prediction \cite{openoccupancy, OCC3D}, BEV semantic segmentation \cite{PON, LSS, VPN, PYVA}, and End-to-end autonomous driving \cite{hu2023_uniad}. It includes 1000 scenes with images from 6 cameras and point clouds from a LiDAR with 32 beams. The camera group has a 360$^\circ$ view which is consistent with LiDAR. This makes it to be a preferable dataset for evaluating algorithms with LiDAR-camera fusion. The scenes are officially split into 700/150/150 scenes for training/validation/testing. There are up to 1.4M annotated 3D bounding boxes for 10 classes: car, truck, bus, trailer, construction vehicle, pedestrian, motorcycle, bicycle, barrier, and traffic cone.

\paragraph{Evaluation Metrics} For 3D object detection, we report the official predefined metrics: mean Average Precision (mAP), Average Translation Error (ATE), Average Scale Error (ASE), Average Orientation Error (AOE), Average Velocity Error (AVE), Average Attribute Error (AAE), and NuScenes Detection Score (NDS). The mAP is analogous to that in 2D object detection \cite{COCO} for measuring the precision and recall, but defined based on the match by 2D center distance on the ground plane instead of the Intersection over Union (IOU) \cite{nuScenes}. NDS is the composite of the other indicators for comprehensively judging the detection capacity. The remaining metrics are designed for calculating the positive results' precision on the corresponding aspects (\emph{e.g.}, translation, scale, orientation, velocity, and attribute).

\paragraph{Predicting Pipeline}
As illustrated in Tab.~\ref{tab:network}, we follow two classical 3D object detection paradigms BEVDet-R50 \cite{BEVDet} and CenterPoint \cite{CenterPoint} to build the image branch and the LiDAR branch separately for ablation study. Besides, we also provide some recommended configurations with superior trade-offs between inference latency and accuracy.

\paragraph{Training and Evaluating}
The DAL models are trained with a batch size of 64 on 16 3090 GPUs. As listed in Tab.~\ref{tab:trainingpipeline}, different from most existing methods that require multiple pre-trained stages and complicated learning rate strategies, DAL loads the pre-trained weight from the ImageNet classification task only and trains the hold pipeline for a total of 20 epochs with CBGS \cite{CBGS}. DAL shares the same simple learning rate policy as CenterPoint \cite{CenterPoint}. Specifically, the learning rate is adjusted by following the cycle learning rate policy \cite{cycle} with an initial value of $2.0 \times 10^{-4}$. During evaluation, we report the performance of a single model without test time augmentation. Inference speeds are all tested on a single 3090 GPU by default. BEVPoolV2 \cite{huang2022bevpoolv2} is used for accelerating the view transformation algorithm LSS.

\begin{table*}
\footnotesize
\begin{center}
\begin{tabular}{c|l|c|c|c|cccccccc}

\hline
Config       &Pipeline   & Aux.      & Resize        &Vel.           & mATE      & mASE      & mAOE      & mAVE      & mAAE      &mAP    & NDS   \\
\hline

A            &BEVFusion-L&           &0.36-0.55      &               & 29.11     & \textbf{24.81}     & 31.01     & 24.55     & 18.91     &63.67  &69.00 \\
B            &BEVFusion  &           &0.36-0.55      &               & 29.08     & 25.22     & 32.05     & 25.64     & 18.82     &63.59  &68.71 \\
C            &BEVFusion  &\checkmark &0.36-0.55      &               & 28.90     & 25.25     & 29.53     & 25.30     & 18.33     &63.45  &68.99 \\
D            &BEVFusion  &\checkmark &0.36-0.88      &               & \textbf{28.25}     & 25.55     & 30.97     & 26.34     & 19.24     &68.00  &70.97 \\
E            &BEVFusion  &\checkmark &0.36-0.88      &\checkmark     & 28.81     & 25.31     & 31.21     & \textbf{19.26}     & \textbf{17.99}     &67.87  &71.67 \\
\hline
F            &DAL        &\checkmark &0.36-0.55      &               & 28.99     & 25.09     & \textbf{29.49}     & 24.13     & 17.96     &64.16  &69.52 \\
G            &DAL        &\checkmark &0.36-0.88      &               & 29.09     & 25.36     & 32.25     & 25.80     & 19.20     &68.07  &70.87 \\
H            &DAL        &\checkmark &0.36-0.88      &\checkmark     & 28.59     & 25.07     & 30.22     & 19.31     & 18.56     &\textbf{68.50}  &\textbf{71.94} \\
\hline
\end{tabular}
\end{center}
\vspace{-0.5cm}
\caption{Ablation study of some key design. All configurations are trained with the same pipeline as DAL in Tab.~\ref{tab:trainingpipeline}. Baseline BEVFusion is a modified version as listed in Tab.~\ref{tab:network}. Aux. means the auxiliary classification task. Resize denotes the range of scale in image resize augmentation with respect to the original image size. Vel. denotes the velocity augmentation. '-L' denotes using the LiDAR modality only. }
\label{tab:ablation}
\end{table*}

\subsection{Benchmark Results}
\paragraph{Results on the nuScenes \texttt{val} set.}
As listed in Tab.~\ref{tab:val} and illustrated in Fig.~\ref{fig:fps-nds}, the proposed DAL paradigm not only substantially pushes the performance boundary, but also provides a better trade-off between speed and accuracy. Configuration DAL-Large scores 71.5 mAP and 74.0 NDS surpassing the best existing record by a large margin of +1.0 mAP and +0.7 NDS. With such high accuracy, DAL-Large still runs at an inference speed of 6.10 FPS. Another recommended configuration DAL-Base runs at a similar inference speed as the fastest method CMT-R50 \cite{CMT}. Its accuracy surpasses CMT-R50 by a large margin of 2.1 mAP and 2.6 NDS. With a similar accuracy as CMT-R50, DAL-Tiny offers an acceleration of 54$\%$.

\paragraph{Results on the nuScenes \texttt{test} set.}
We report the performance of configuration DAL-Large on the nuScenes \texttt{test} set without model ensemble and test time augmentation. DAL outshines all other approaches in terms of NDS 74.8.

\subsection{Ablation Study}
We use experiments illustrated in Tab.~\ref{tab:ablation} to reveal the impact of the key designs. We take a modified version of BEVFusion as a baseline as detailed in Tab.~\ref{tab:network}. As a reference configuration, configuration A in Tab.~\ref{tab:ablation} uses the LiDAR modality only. It cores 63.67 mAP and 69.00 NDS. In Tab.~\ref{tab:ablation} configuration B, we train the BEVFusion paradigm as DAL. No superiority can be observed in the performance of this configuration (\texttt{i.e.} 63.59 mAP and 68.71 NDS) on the baseline. This means that BEVFusion relies on a complicated pre-trained strategy to take advantage of the image modality. Directly loading pre-trained weight from the ImageNet classification task and training the hold pipeline is not feasible for BEVFusion.

With the DAL predicting pipeline and the auxiliary classification task, configuration F in Tab.~\ref{tab:ablation} scores 64.16 mAP and 69.52 NDS mildly surpassing the LiDAR-only baseline (A) by +0.49 mAP and +0.52 NDS. This indicates that the DAL pipeline built upon the 'Detecting As Labeling' concept is a feasible solution to take advantage of the visual modality even with a simple training pipeline. By applying a large range of image resize augmentation in Tab.~\ref{tab:ablation} configuration G, the accuracy is improved by a large margin of +3.91 mAP and +1.35 NDS to 68.07 mAP and 70.87 NDS. Freeing the image branch from regression tasks and making a large range of image resize augmentation feasible is another key factor of the DAL paradigm. In addition, velocity augmentation applied in Tab.~\ref{tab:ablation} configuration H offers extra performance boosting of 0.47 mAP and 1.07 NDS. A more balanced velocity distribution in the training set significantly decreases the predicted error by 25$\%$.

Last but not least, we use configurations C, D, and E in Tab.~\ref{tab:ablation} to reveal something interesting. First of all, an extra auxiliary classification task in configuration C does not offer an improvement on a BEVFusion-like predicting pipeline. We conjecture that fusing the feature before the dense BEV encoder is too early to maintain a relatively independent judgment from image cues. Second, by using a large range of resizing augmentation, configuration D with a BEVFusion-like predicting pipeline performs close to the DAL-like one in configuration G. A large range of resizing augmentation destroys the connection between the image cues and regression task prediction and forces the model to concentrate on the point cloud cues instead. So explicitly removing image cues from regression task prediction like the DAL-like predicting pipeline is not the only way to achieve the goal of 'Detecting As Labeling'. Instead of a novel predicting pipeline, what is more valuable to this problem is the cognitive change by the fundamental theory updating. With additional velocity augmentation in Configuration E, a BEVFusion-like predicting pipeline scores 67.87 mAP and 71.67 NDS, slightly lagging behind the DAL-like one in Configuration H. We conjecture that the Velocity augmentation challenges the regression task prediction with point cloud cues, which forces the model to utilize the image cues. So we prefer to use a DAL-like predicting pipeline to prevent the model from overfitting the image cues in regression task prediction under some extreme situations (\texttt{e.g.} strong augmentation on the point cloud modality, image branch capacity enlargement).

\begin{table}[t]
\footnotesize
\begin{center}
\begin{tabular}{c|c|c|c|c}
\hline
Factor                              & Value              & FPS   & mAP   & NDS\\
\hline
\multirow{5}{*}{Image Resolution}   & $160\times448$     & 15.86 &67.08 &71.28\\
                                    & $192\times544$     & 15.40 &67.59 &71.51\\
                                    & $256\times704$     & 12.88 &68.23 &71.94\\
                                    & $384\times1056$    & 10.23 &68.69 &72.13\\
                                    & $512\times1408$    & 7.80  &68.77 &71.98\\
\hline
\multirow{3}{*}{Image Backbone}   & R18     & 15.87 &67.49 &71.47\\
                                    & R50     & 12.88 &68.23 &71.94\\
                                    & R101    & 10.99 & 67.86 &71.62\\
\hline
\multirow{3}{*}{\makecell{VoxelNet Resolution\\(Meter)}}
& 0.100    & 12.88  &68.23  &71.94\\
& 0.075    &  11.69      &69.33  &72.64\\
& 0.050    &   9.70     &69.23  &72.18\\

\hline
\multirow{4}{*}{\makecell{VoxelNet \\Sparse Encoder \\(Base Channels)}}
& 16    & 12.88 &68.23 &71.94\\
& 24    & 11.93  &68.24 &72.01\\
& 32    & 11.12 &69.36  &72.66\\
& 48    & 8.67 &67.82  &72.02\\

\hline
\multirow{3}{*}{\makecell{VoxelNet \\Dense Encoder (Blocks)}}
& 3    & 13.29 &67.54  &71.27\\
& 5    & 12.88 &68.23 &71.94\\
& 8    & 12.73 & 67.84 &71.57\\
\hline
\multirow{3}{*}{\makecell{(0.075) VoxelNet \\Dense Encoder (Blocks)}}
& 5    & 11.69 &69.33  &72.64\\
& 8    &  11.37 & 69.84 &73.06\\
& 12    & 10.94  & 69.52 &72.78\\
\hline
\end{tabular}
\end{center}
\vspace{-0.5cm}
\caption{Ablation study on the pivotal construction details.}
\label{tab:abl-network}
\vspace{-0.2cm}
\end{table}

\paragraph{Predicting Pipeline}
In Tab.~\ref{tab:abl-network}, we conduct ablation on some key factors of the DAL predicting pipeline from the perspective of accuracy and inference latency. Compared with the default setting R50-256$\times$704, further improving the resolution of the input image or the capacity of the backbone offers finite improvement in accuracy. However, the inference latency increases remarkably. As the camera branch mainly focuses on the classification task, a small backbone with low input resolution is enough for the nuScenes \cite{nuScenes} dataset with only 10 classes and 750 scenes. By contrast, further improving the resolution of the LiDAR branch alone with a larger capacity of the backbones offers remarkable improvement in accuracy. So, it is recommended to incorporate a small image branch with a large LiDAR branch in DAL for a better trade-off between inference latency and accuracy. We follow this to construct some recommended configurations in Tab.~\ref{tab:network}.

\begin{figure}
	\setlength{\abovecaptionskip}{0.cm}
    \begin{center}
        \includegraphics[width=1.0\hsize]{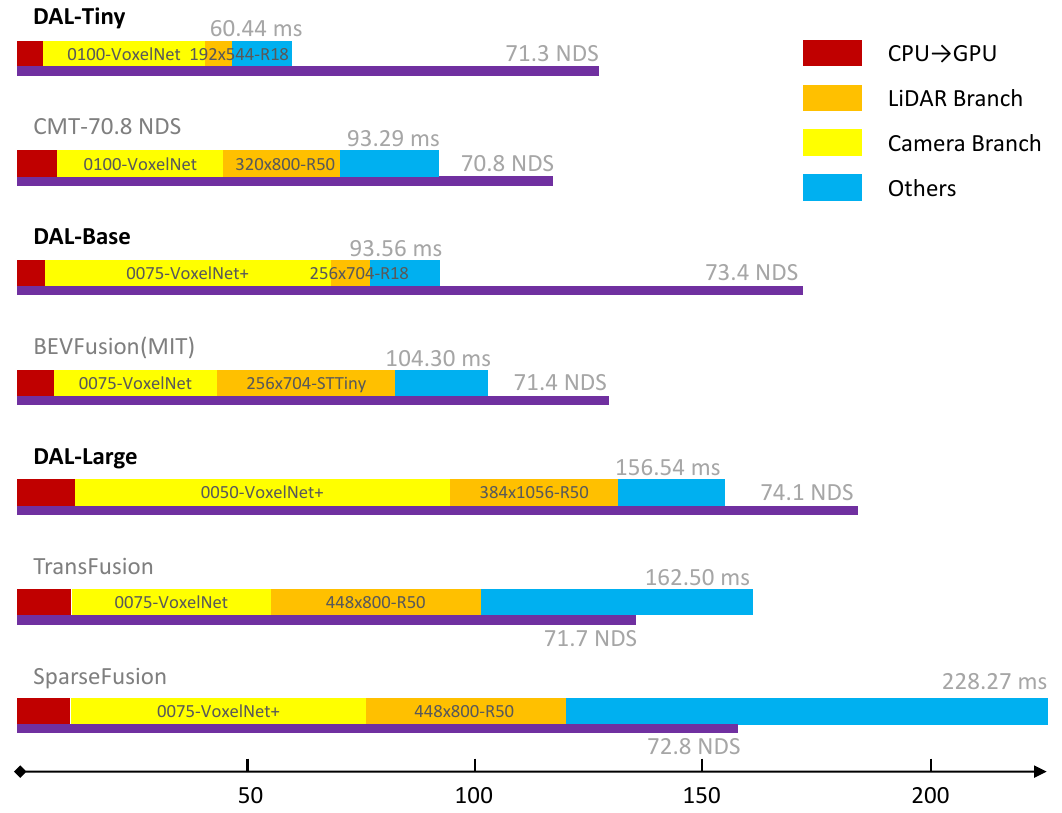}
    \end{center}
   \caption{Inference latency analysis and comparison.}
    \label{fig:latency}
\vspace{-0.2cm}
\end{figure}

\paragraph{Inference Latency Analysis}
In Fig.~\ref{fig:latency}, we illustrate the inference latency of each component in the DAL predicting pipeline and compare it with some leading methods. The inference latency includes the time for transferring data from CPU to GPU, extracting the point cloud feature with the LiDAR branch, extracting the image feature with the camera branch, and executing the remainder processes like multi-modality feature fusion and target prediction. Thanks to the elegant predicting pipeline design in DAL, it spends less time in the 'Other' aspect. Besides, the LiDAR branch in DAL always occupies a high proportion of the inference time. As the LiDAR branch undertakes more missions than the camera branch, the recommended configurations in DAL always use a larger LiDAR branch incorporated with a relatively small camera branch. In contrast, involving the image feature in the regression task prediction, existing methods are always equipped with a large image branch. This degenerates their time efficiency to a certain extent.

\section{Conclusion, Limitation, and Future Work}
\label{sec:CC}

\paragraph{Conclusion} In this paper, we proposed a cutting-edge perspective 'Detecting As Labeling' for 3D object detection with LiDAR-camera fusion. DAL is developed as a template by following this perspective. DAL is an extremely elegant paradigm with a concise predicting pipeline and an easy training process. Though simple in these aspects, it substantially pushes the performance boundary in 3D object detection with LiDAR-camera fusion alone with the best trade-off between speed and accuracy. So it is an excellent milestone for both future work and practical usage.

\paragraph{Limitation and Future Work} Objects beyond the scope of LiDAR have not been considered in DAL. We have tried discriminating this situation by predicting dense heatmaps with the point cloud feature only and comparing them with those predicted with the fuse feature. Then, the regression targets of these instances are predicted with another FFN on the fuse feature instead. However, this modification contributes less to the final accuracy. This is because only the targets with more than 1 LiDAR point will be annotated in nuScenes \cite{nuScenes}. In addition, the scope range is small enough in nuScenes evaluation which ensures sufficient LiDAR points for predicting the regression aspects.

Besides, the simple classification task in the nuScenes dataset limits DAL to apply the advanced image backbone like SwinTransformer \cite{Swin}, DCN \cite{DCN}, and EfficientNet \cite{EfficientNet}. The open-world classification task is far more complicated and thus difficult. Thus, the image branch may take advantage of the advanced image backbones in practice.

Though DAL has an attention-free predicting pipeline, it is just a template to reveal the value of 'Detecting As Labeling'. So we use the most classical algorithms \cite{ResNet, VoxelNet} without applying attention. However, we do not intentionally exclude it from DAL. On the contrary, we regard attention as an appealing mechanism for further developing DAL in many aspects. For example, we can apply advanced DSVT \cite{DSVT} backbone like UniTR \cite{UniTR}, apply attention-based LiDAR-camera fusion like CMT \cite{CMT}, and apply an attention-based sparse detection paradigm like DETR \cite{DETR}.

{
    \small
    \bibliographystyle{ieeenat_fullname}
    \bibliography{main}
}


\end{document}